\documentclass[letterpaper, 10 pt, conference]{ieeeconf}  

\IEEEoverridecommandlockouts                              

\usepackage{bm}
\usepackage{url}
\usepackage{graphicx}
\usepackage{lipsum}
\usepackage{pifont}
\usepackage{amsmath}
\usepackage{amssymb}
\usepackage{nccmath}
\usepackage{comment}
\usepackage{balance}
\usepackage{multirow}
\usepackage{textcomp}
\usepackage{multirow}
\usepackage{subcaption}
\usepackage{booktabs}
\usepackage{mathtools}
\usepackage{algorithm}
\usepackage{dblfloatfix}
\usepackage[table]{xcolor}
\usepackage[noend]{algpseudocode}

\definecolor{verylightgray}{gray}{0.95}
\newcolumntype{g}{>{\columncolor{verylightgray}}c}

\title{\LARGE \bf
Tightly Coupled Range Inertial Odometry and Mapping \\ with Exact Point Cloud Downsampling
}

\author{Kenji Koide$^{1}$, Aoki Takanose$^{1}$, Shuji Oishi$^{1}$, and Masashi Yokozuka$^{1}$
\thanks{*This work was supported in part by JSPS KAKENHI Grant Number 23K16979 and projects, JPNP14004 and JPNP21004, subsidized by the New Energy and Industrial Technology Development Organization (NEDO).}
\thanks{$^{1}$All the authors are with the Department of Information Technology and Human Factors, the National Institute of Advanced Industrial Science and Technology, Tsukuba, Ibaraki, Japan, {\tt\small k.koide@aist.go.jp}}%
}

\begin{document}

\maketitle
\thispagestyle{empty}
\pagestyle{empty}

\setlength\floatsep{6pt}
\setlength\textfloatsep{6pt}

\begin{abstract}

In this work, to facilitate the real-time processing of multi-scan registration error minimization on factor graphs, we devise a point cloud downsampling algorithm based on coreset extraction. This algorithm extracts a subset of the residuals of input points such that the subset yields exactly the same quadratic error function as that of the original set for a given pose. This enables a significant reduction in the number of residuals to be evaluated without approximation errors at the sampling point. Using this algorithm, we devise a complete SLAM framework that consists of odometry estimation based on sliding window optimization and global trajectory optimization based on registration error minimization over the entire map, both of which can run in real time on a standard CPU. The experimental results demonstrate that the proposed framework outperforms state-of-the-art CPU-based SLAM frameworks without the use of GPU acceleration.

\end{abstract}

\section{Introduction}

Point cloud SLAM algorithms that directly compute and minimize point cloud registration errors on factor graphs have been gaining attention due to their precision and robustness. Methods such as odometry estimation with sliding window optimization \cite{Nguyen_2023}, global trajectory optimization via global registration error minimization \cite{Koide2021}, and LiDAR-bundle adjustment \cite{Liu2021} excel in optimizing sensor poses to maximize the consistency between multiple point clouds. They offer more accurate and reliable estimations compared to those obtained using traditional approaches such as filtering-based odometry estimation \cite{Xu2022} and global trajectory optimization using relative pose constraints \cite{behley2018efficient}. However, those methods are computationally intensive, as they simultaneously optimize multiple sensor poses with registration error factors involving a vast number of points. This makes real-time processing challenging without extensive approximations or hardware accelerators such as GPUs \cite{Nguyen_2023,Koide2021,Liu2023}.

In our previous work, we introduced GLIM, a SLAM framework that utilizes GPU-accelerated registration error factors \cite{Koide_2024}. Its odometry estimation and global trajectory optimization rely on factor graphs that directly minimize point cloud registration errors across multiple frames, eliminating the need for conventional relative pose factors (i.e., {\it gtsam::BetweenFactor{\textless}Pose3{\textgreater}}). These registration-error-minimization-based algorithms demonstrated significant advantages, including extreme robustness against point cloud degeneration and interruptions as well as the ability to close loops even with minimal frame overlap. However, the reliance on GPU acceleration limits their broad applicability.

\begin{figure}[tb]
  \centering
  
  \begin{minipage}[tb]{\linewidth}
  \centering
  \includegraphics[width=\linewidth]{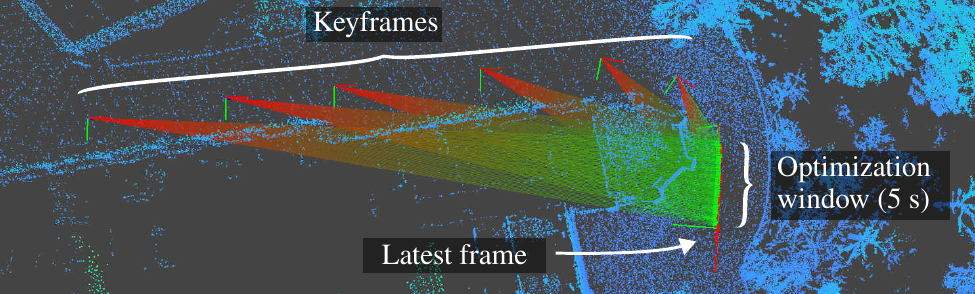} \vspace{-5mm}
  \subcaption{Odometry estimation factor graph}
  \end{minipage}
  \begin{minipage}[tb]{\linewidth}
  \centering
  \includegraphics[width=\linewidth]{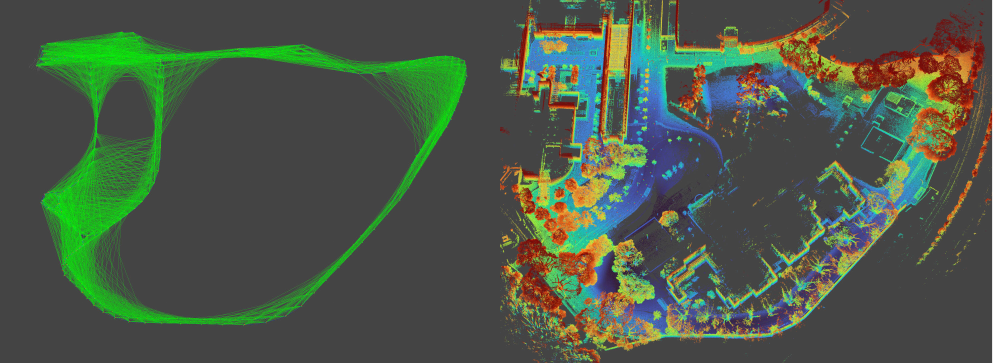} \vspace{-5mm}
  \subcaption{Global trajectory optimization factor graph}
  \end{minipage}
  \caption{Factor graphs for odometry estimation and global trajectory optimization. The proposed framework extensively uses registration error factors to directly minimize registration errors across multiple point clouds. The exact downsampling algorithm drastically reduces the linearization cost of registration error factors and enables the real-time processing of these dense factor graphs on standard CPUs.}
  \label{fig:graphs}
\end{figure}

In the present work, we aim to enable the real-time optimization of registration error minimization factor graphs using only a standard CPU\footnote{See the project page for supplementary videos: \url{https://staff.aist.go.jp/k.koide/projects/icra2025_es/}.}. To this end, we leverage a point cloud downsampling algorithm based on efficient coreset extraction \cite{Koide_2023}. This method extracts a subset of the residuals of input points, ensuring that the resulting subset preserves the exact quadratic error function of the original set for a given pose. This approach significantly reduces the number of residuals to be evaluated without introducing approximation errors at the sampling point. We also introduce a strategy to mitigate the computational cost of the downsampling by deferring the coreset extraction until it is required. By integrating this sampling algorithm, we develop a complete SLAM framework with registration error minimization factor graphs that can operate in real time on standard CPUs.

The contributions of this work are as follows:
\begin{itemize}
  \item We extend the exact point cloud downsampling algorithm~\cite{Koide_2023} by incorporating point correspondence updating to better capture the nonlinearity of the registration error function. We also introduce a factor linearization strategy that defers the execution of exact downsampling to reduce the computational overhead of the linearization process.
  \item Leveraging exact point cloud downsampling, we develop a complete SLAM framework that extensively uses registration error factors for both odometry estimation and global trajectory optimization, as shown in Fig.~\ref{fig:graphs}. Although the graph structures were originally designed for GPU processing~\cite{Koide_2024}, the exact downsampling algorithm enables them to operate in real time on a standard CPU.
\end{itemize}

\section{Related Work}

\subsection{Odometry Estimation}

Many existing LiDAR odometry algorithms rely on state filtering, which optimizes only the current sensor state while marginalizing past states as new observations arrive \cite{Xu2022,Chen_2023}. The process is typically combined with scan-to-model matching, where past point clouds are accumulated into a single model point cloud (or local map), and the current sensor scan is aligned with this model \cite{zhang2014loam}. Although this causal estimation approach is efficient and accurate in feature-rich environments, it struggles to propagate the uncertainty of past observations and sensor states, resulting in difficulties in scenarios with point cloud degeneration and interruptions.

To enhance robustness, LiDAR odometry algorithms based on sliding window optimization have been introduced~\cite{Nguyen_2023, Koide_2024, Zuo_2020, Li_2021}. Unlike filtering-based approaches, these algorithms optimize both current and past sensor states within a sliding window and allow the correction of estimation drift by propagating uncertainties backward in time. This results in robustness to point cloud degeneration and rapid motion. However, the need to continuously optimize multiple sensor states requires significant computational resources, making real-time processing challenging \cite{Nguyen_2023}. Consequently, many of these methods rely on extensive downsampling \cite{Li_2021} or feature extraction \cite{Zuo_2020} to reduce the computational demands.

\subsection{Global Trajectory Optimization}

Pose graph optimization constructs a factor graph based on relative pose constraints and optimizes the global trajectory by minimizing errors in the pose space. This method has been widely adopted due to its computational efficiency \cite{Nguyen_2023, behley2018efficient}. However, its accuracy is often constrained by the challenge of accurately representing the uncertainty of scan matching results using a Gaussian distribution \cite{Koide2021}.

Integrating point cloud registration errors into global trajectory optimization can improve the accuracy of trajectory estimation \cite{Koide_2024, Reijgwart2020}. However, computing these errors across an entire map is computationally expensive. Existing approaches rely on significant downsampling \cite{Yuan_2022} or GPU computation \cite{Koide_2024} to manage the high computational load.

\section{Methodology}

\begin{figure}[tb]
  \centering
  \includegraphics[width=\linewidth]{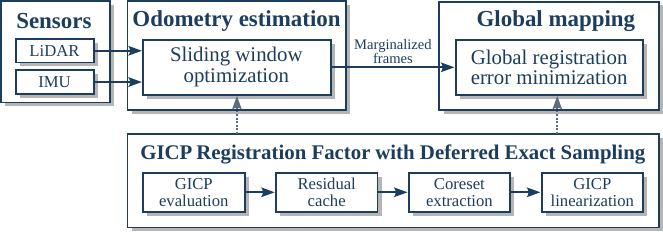}
  \caption{Odometry estimation module, which estimates sensor ego-motion using sliding window factor graph optimization, and global mapping module, which constructs factor graph to directly minimize matching cost errors across entire map. Both modules utilize the GICP scan matching factor accelerated with the exact point cloud downsampling algorithm.}
  \label{fig:system}
\end{figure}

Fig.~\ref{fig:system} shows an overview of the proposed framework, which comprises odometry estimation and global trajectory estimation modules. Both modules are built with two key building blocks, namely the generalized iterative closest point (GICP) registration error factor accelerated with exact point cloud downsampling and voxel-based fast overlap estimation with occupancy bit chunks. We introduce these building blocks in Secs.~\ref{sec:gicp} and \ref{sec:overlap} and then describe the proposed SLAM framework in Secs.~\ref{sec:odom} and \ref{sec:mapping}.

\subsection{Registration Error Factor Accelerated with Exact Point Cloud Downsampling}
\label{sec:gicp}

{\bf Registration error function: } To constrain the relative pose between two point clouds $\mathcal{P}_i$ and $\mathcal{P}_j$, we use the distribution-to-distribution distance metric in GICP \cite{segal2009generalized}. GICP models each point ${\bm p}_k \in \mathcal{P}_j$ as a Gaussian distribution ${\bm p}_k = ({{\bm \mu}_k}, {{\bm \Sigma}_k})$, which represents the local surface shape, and computes the distance between point ${\bm p}_k$ and its corresponding nearest point ${\bm p}'_k = ({{\bm \mu}'_k}, {{\bm \Sigma}'_k})$ in the other point cloud as follows:
\begin{align}
f^\text{GICP} (\mathcal{P}_i, \mathcal{P}_j, {\bm T}_{ij}) &= \sum_{{\bm p}_k \in \mathcal{P}_j} \| f^\text{D2D} ({\bm p}'_k, {\bm p}_k, {\bm T}_{ij}) \|^2, \\
f^\text{D2D} ({\bm p}'_k, {\bm p}_k, {\bm T}_{ij}) &= {\bm \Phi}^\top_k {\bm d}_k, \\
{\bm d}_k &= {{\bm \mu}}'_k - {{\bm T}_{ij}} {{\bm \mu}_k}, \\
{\bm \Phi}_k{\bm \Phi}_k^\top &= ({{\bm \Sigma}'_k} + {{\bm T}_{ij}} {{\bm \Sigma}_k} {{\bm T}^\top_{ij}})^{-1},
\end{align}
where ${{\bm T}_{ij}} = {\bm T}_i^{-1} {\bm T}_j$ is the relative pose between $\mathcal{P}_i$ and $\mathcal{P}_j$. The decomposition ${\bm \Phi}_k{\bm \Phi}_k^\top$ of the information matrix can be efficiently obtained using Cholesky decomposition.

In the Gauss-Newton optimization, the residual function $f^\text{GICP}$ is linearized at the current estimate $\breve {\bm T}_{ij}$ to form a quadratic error factor:
\begin{align}
f^\text{GICP} (\mathcal{P}_i, \mathcal{P}_j, \breve {\bm T}_{ij} \boxplus \Delta{\bm x}) \approx \Delta {\bm x}^\top {\bm H} \Delta {\bm x} + 2 {\bm b}^\top \Delta {\bm x} + c,
\end{align}
where ${\bm H} = {\bm J}^\top {\bm J},~{\bm b} = {\bm J}^\top {\bm e},~c = {\bm e}^\top {\bm e},~{\bm J} = \frac{\partial {\bm e}}{\partial {\bm T}}$, and ${\bm e}$ is a stack of residuals given by $f^\text{D2D}$.

{\bf Exact point cloud downsampling: } The linearization of $f^\text{GICP}$ is computationally intensive because it requires the evaluation of residuals for all points in $\mathcal{P}_j$. A common approach to mitigate this computation burden is to decrease the number of points using random sampling \cite{Reijgwart2020} or geometry-aware feature selection \cite{Yi_2024, Duan_2022, Jiao_2021}. However, the accuracy of these sampling methods largely depends on the number of sampled points and typically requires sampling several thousand points to maintain sufficient accuracy.

To drastically reduce the number of points (e.g., tens to hundreds) while retaining accuracy, we adopt the exact point cloud downsampling algorithm \cite{Koide_2023}, which is based on the concept of coresets in computational geometry  \cite{NEURIPS2019_475fbefa}. 

An exact coreset $\mathcal{X}'$ is a subset of input data $\mathcal{X}$ selected such that the result of an algorithm $f$ on the coreset becomes the same as that on the original set: $f(\mathcal{X}') = f(\mathcal{X})$, where $\mathcal{X}' \subset \mathcal{X}$. Our downsampling algorithm extracts an exact coreset of the residuals of input points, ensuring that the original quadratic error function is precisely recovered. Specifically, given a relative pose (referred to as the sampling point) $\tilde {\bm T}_{ij}$, it extracts a subset of residuals $\tilde {\bm e} \subset {\bm e}$ and corresponding weights $\tilde {\bm w}$ such that the weighted subset yields exactly the same quadratic error function parameters $\tilde {\bm H}$, $\tilde {\bm b}$, and $\tilde c$ as those of the original set (${\bm H}, {\bm b}$, and $c$) at $\tilde{\bm T}_{ij}$:
$\tilde {\bm H} = \tilde {\bm J}^\top \tilde {\bm W} \tilde {\bm J} = {\bm H},$
$\tilde {\bm b} = \tilde {\bm J}^\top \tilde {\bm W} \tilde {\bm e} = {\bm b},$
$\tilde {c}     = \tilde {\bm e} \tilde {\bm W} \tilde {\bm e} = c,$
where $\tilde {\bm J} = \frac{\partial \tilde {\bm e}}{\partial {\bm T}_{ij}}$ and $\tilde {\bm W} = \text{diag}(\tilde {\bm w})$. Due to space limitations, we refer the reader to \cite{Koide_2023,NEURIPS2019_475fbefa} for the detailed process of finding such a coreset\footnote{The coreset extraction algorithm is available at \url{https://github.com/koide3/caratheodory2}.}.

When re-linearizing $f^\text{GICP}$, we evaluate only the selected subset $\tilde {\bm e}$ and compose a quadratic error factor. Since the subset $\tilde {\bm e}$ yields the same quadratic error function parameters as those for the original set ${\bm e}$, no approximation errors are introduced at the sampling point $\tilde {\bm T}_{ij}$. As the subset $\tilde {\bm e}$ can be much smaller than the original set (about 0.5\% to 5\% of ${\bm e}$), re-linearization is significantly faster compared to the original set. Unlike prior work \cite{Koide_2023}, where $f^\text{GICP}$ was re-linearized without updating point correspondences, our approach re-linearizes $f^\text{GICP}$ with the coreset and updates point correspondences. While the coreset does not guarantee approximation accuracy for nonlinearity, in practice, it enables an accurate approximation of the nonlinear objective function, as demonstrated in Sec.~\ref{sec:ex_approx}. 

\begin{figure}[tb]
  \centering
  \includegraphics[width=0.8\linewidth]{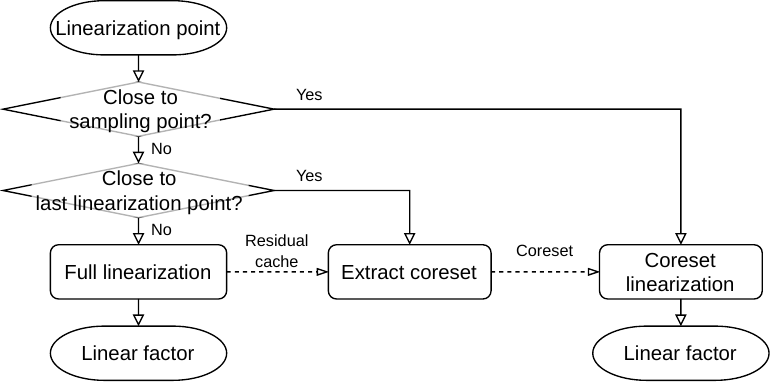}
  \caption{Flowchart of exact point cloud downsampling with deferred sampling strategy.}
  \label{fig:deffered}
\end{figure}

{\bf Deferred sampling strategy: } Although the extracted coreset provides good approximation accuracy around the sampling point $\tilde {\bm T}_{ij}$, the approximation error grows as the current estimate $\breve {\bm T}_{ij}$ deviates further from $\tilde {\bm T}_{ij}$. We thus re-extract a new coreset when the translation or rotation displacement between the current estimate $\breve {\bm T}_{ij}$ and the sampling point $\tilde {\bm T}_{ij}$ exceeds its threshold (e.g., 1.0 m or 1.0\textdegree, respectively).

However, naively re-extracting the coreset every time $f^\text{GICP}$ is linearized would introduce overhead, especially during the initial optimization iterations when sensor pose estimates tend to change significantly, causing the newly extracted coresets to be quickly discarded in the next linearization. To mitigate this, we implement a deferred sampling strategy, as shown in Fig.~\ref{fig:deffered}

In this strategy, the GICP factor is first linearized at the current estimate $\breve {\bm T}^0_{ij}$ with all input points, and residuals ${\bm e}^0$ and their Jacobians ${\bm J}^0$ are obtained. We store ${\bm e}^0$ and ${\bm J}^0$ in cache memory and create a quadratic factor with the full residuals ${\bm e}^0$. During the next linearization, if the displacement between the current and previous linearization points, $(\breve {\bm T}^0_{ij})^{-1} \breve {\bm T}^1_{ij}$, is below its threshold (e.g., 0.25 m or 0.25\textdegree), we perform exact downsampling on the cached residuals ${\bm e}^0$ and Jacobians ${\bm J}^0$ to extract the coreset $\tilde {\bm e}$. This coreset is then re-evaluated at the current linearization point $\breve {\bm T}_{ij}^1$ to form a new quadratic factor. As long as the current estimate $\breve {\bm T}_{ij}^t$ remains close to the sampling point $\tilde {\bm T}_{ij} = \breve {\bm T}^0_{ij}$, the coreset is reused to re-linearize the objective function $f^\text{GICP}$.

Since the displacements of sensor pose estimates tend to converge as the optimization proceeds, this deferred evaluation strategy significantly reduces unnecessary executions of exact downsampling.

\subsection{Fast Overlap Estimation}
\label{sec:overlap}

We use a voxelmap-based overlap metric to manage the localization and mapping processes. To enhance processing speed, we employ a spatial voxel hashing algorithm \cite{teschner2003optimized} similar to those in VoxelMap \cite{Yuan_2022} and Faster-LIO \cite{Bai2022}. Additionally, we introduce a fast occupancy grid based on occupancy bit chunks to further improve efficiency. Inspired by the VDB structure in \cite{Museth_2013}, we pack the binary occupancy states of $8 \times 8 \times 8$ voxels into a 512-bit chunk and store the occupancy chunks in a flat hash table. Hash collisions are resolved by open addressing. Unlike implementations \cite{Yuan_2022,Bai2022} that rely on closed addressing (e.g., with std::unordered\_set), our implementation stores all voxel data in a compact contiguous memory region, making it more cache-friendly. We compute the overlap ratio by counting the number of source points that fall within occupied target voxels. We observed that this method yields up to a 2.5x processing speed improvement compared to conventional implementations \cite{Yuan_2022,Bai2022}. Note that we use this occupancy grid solely for overlap evaluation. The nearest neighbor search in GICP factors is performed using an exact nearest neighbor search, such as KdTree.


\subsection{Odometry Estimation}
\label{sec:odom}

\begin{figure}[tb]
  \centering
  \includegraphics[width=0.9\linewidth]{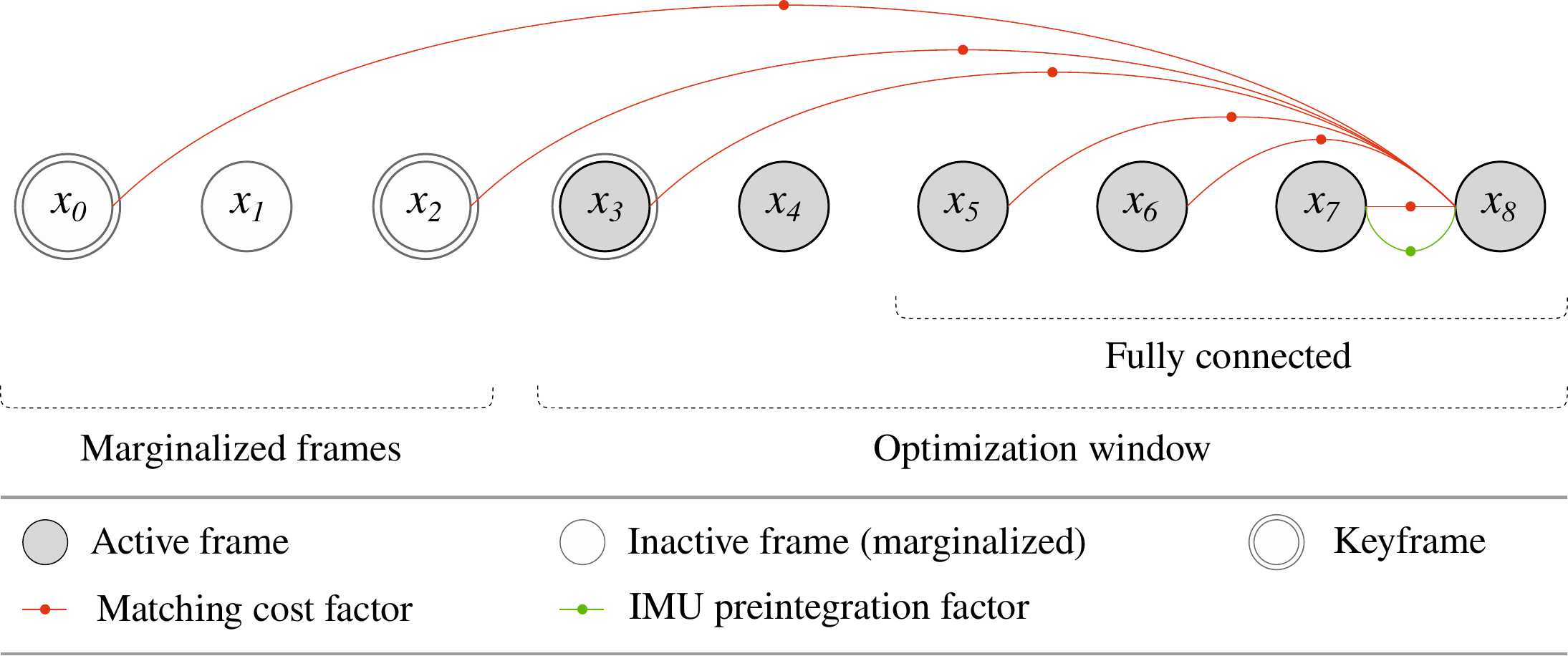}
  \caption[cap]{Factor graph for odometry estimation\footnotemark.}
  \label{fig:graph_odom}
\end{figure}

Following \cite{Koide_2024}, we employ keyframe-based odometry estimation with sliding window factor graph optimization, as shown in Fig.~\ref{fig:graph_odom}. Let ${\bm x}_t = [{\bm T}_t, {\bm v}_t, {\bm b}_t]$ be the sensor state at time $t$, where ${\bm T}_t \in SE(3)$ is the sensor pose, ${\bm v}_t \in \mathbb{R}^3$ is the velocity, and ${\bm b}_t \in \mathbb{R}^6$ is the IMU bias.

We optimize the sensor states within an optimization window $\mathcal{X}^w$ that contains frames from the past 5 seconds. To bound the computational cost, frames that leave the window are marginalized from the graph. Every time a new sensor frame is inserted, we create a preintegrated IMU factor \cite{Forster_2015} between the current and previous frames. We also create GICP factors between the new frame and the preceding $N^\text{pre}$ frames (e.g., three frames) to improve robustness to rapid sensor motion. To reduce estimation drift, GICP factors are also created between the latest frame and keyframes $\mathcal{X}^k$, which is a set of past frames selected such that they are well distributed in space (i.e., with maximal distance between them) while having sufficient overlap with the latest frame.

\footnotetext{Reprinted from \cite{Koide_2024} with permission from Elsevier.}

The objective function is summarized as follows:
\begin{align}
g^\text{LIO} ( \mathcal{X}^w ) &= \sum_{{\bm x}_j \in \mathcal{X}^w} \sum_{{\bm x}_i \in \mathcal{X}_j^p \cup \mathcal{X}^k} f^\text{GICP} ( \mathcal{P}_i, \mathcal{P}_j, {\bm T}_i, {\bm T}_j  ) \\
  & + \sum_{{\bm x}_i \in \mathcal{X}^w} f^\text{IMU} ( {\bm x}_{i - 1}, {\bm x}_i ) + f^\text{MG}(\mathcal{X}^w),
\end{align}
where $\mathcal{X}^p_j = \{ {\bm x}_{j - N^\text{pre}}, \ldots, {\bm x}_{j - 1} \}$ is the preceding frames of ${\bm x}_j$, $\mathcal{X}^k$ is the keyframes,  $f^\text{IMU}$ is the IMU error factor, and $f^\text{MG}$ is the error term to compensate for marginalized variables and factors. We use the iSAM2 optimizer \cite{Kaess_2011} implemented in GTSAM \cite{gtsam} for efficient optimization and marginalization.

A key difference from \cite{Koide_2024} is that we use the GICP factor with exact nearest neighbor search (i.e., KdTree) instead of voxel-based nearest neighbor search. The use of exact nearest neighbor search improves the convergence of the optimization process. While this results in a high computational cost, exact downsampling enables real-time processing on a CPU and makes GPU computation unnecessary.

\subsection{Global Trajectory Optimization}
\label{sec:mapping}

Frames marginalized from the odometry estimation graph are concatenated into a single point cloud at a certain interval (e.g., every 50 frames) to form a submap. The global mapping module then constructs a factor graph to optimize the submap poses to achieve a globally consistent sensor trajectory. As in \cite{Koide_2024}, we create a GICP factor between every pair of submaps with an overlap that exceeds a threshold (e.g., 15\%), resulting in a highly dense factor graph, as shown in Fig.~\ref{fig:graphs} (b). The objective function is thus defined as
\begin{align}
g^\text{GM}(\mathcal{X}^g) = \sum_{ ({\bm T}^i, {\bm T}^j) \in \mathcal{X}^o } f^\text{GICP} ( \mathcal{S}^i, \mathcal{S}^j, {\bm T}^i, {\bm T}^j ),
\end{align} 
where $\mathcal{X}^g$ is the set of all submap poses, $\mathcal{X}^o$ is all pairs of submaps with an overlap, and $\mathcal{S}^i$ and ${\bm T}^i$ are the point cloud and the pose of submap $i$, respectively. The optimization is incrementally performed with the iSAM2 optimizer \cite{Kaess_2011}.

\section{Experiments}


\subsection{Approximation Accuracy of Exact Downsampling}
\label{sec:ex_approx}

\begin{figure}[tb]
  \centering
  \includegraphics[width=\linewidth]{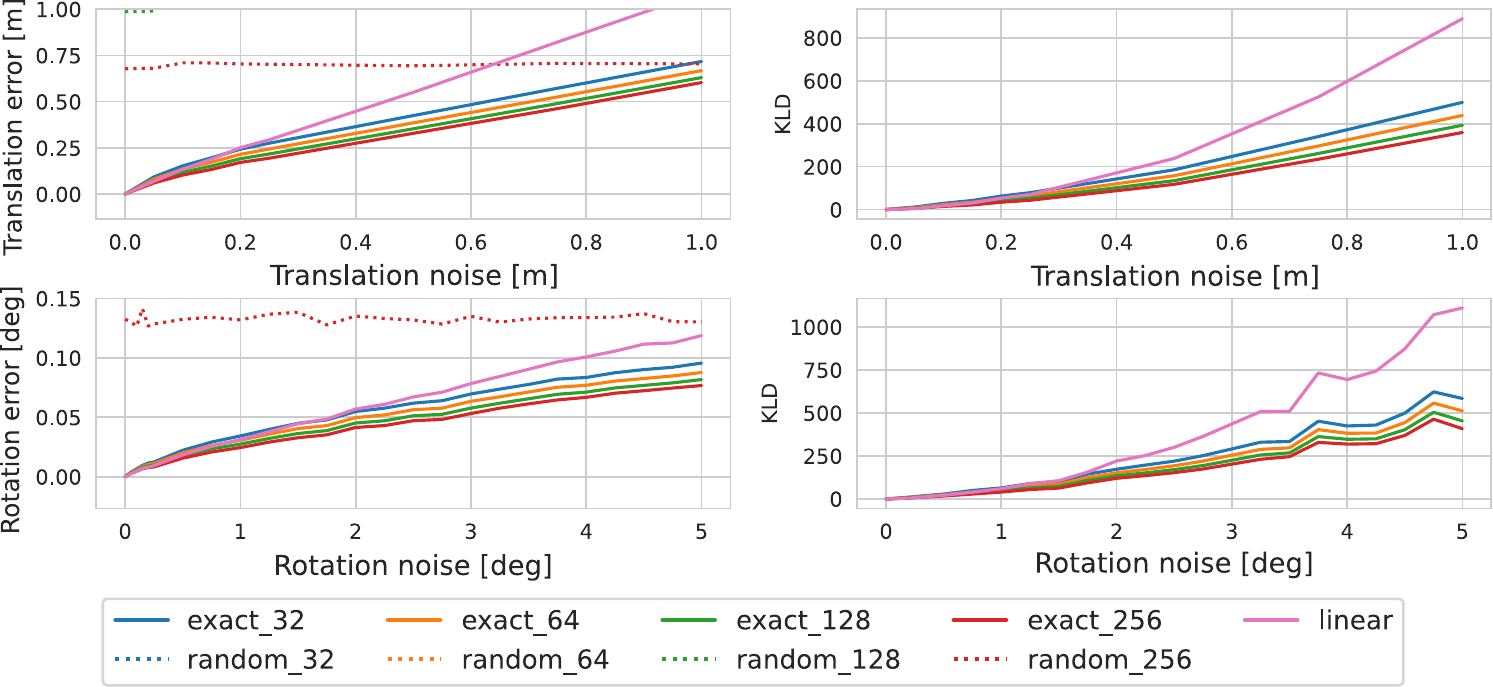}
  \caption{Registration error function approximation errors. Exact downsampling showed zero errors at the sampling point (when noise = 0) and consistently showed errors smaller than those of the quadratic approximation (i.e., linear factor at the sampling point). Note that most of the random sampling results are not visible in this figure due to excessively large errors. The numerical values in the legend refer to the numbers of points sampled.}
  \label{fig:approx_mean}
\end{figure}


{\bf Experimental setting: }
We evaluated the approximation accuracy of exact point cloud downsampling on the KITTI dataset \cite{geiger2013vision}. We created pairs of consecutive frames from the first 500 frames of the KITTI 00 sequence and extracted an exact coreset for each pair. We then compared the linearization results of the registration error on the coreset and those on the original points under pose perturbation. The approximation error was measured using two metrics: 1) KL divergence (KLD) between the covariance matrices (i.e., $\text{KLD}({\bm H}^{-1}, \tilde{\bm H}^{-1})$ ), 2) translation and rotation errors between the mean vectors (i.e., ${\bm \mu} = {\bm H}^{-1} {\bm b}$ and $\tilde {\bm \mu} = \tilde {\bm H}^{-1} \tilde {\bm b}$). For each frame pair, we repeated the evaluation 100 times, varying the magnitude of the displacement from the sampling point. As baselines, we also evaluated random sampling and quadratic approximation (i.e., linearized factor at the sampling point). The number of sampled points varied from 32 to 256. Note that random sampling often yields corrupted linearized factors with such a small number of sampled points.

{\bf Experimental results: }
Fig.~\ref{fig:approx_mean} shows the evaluation results. Exact downsampling demonstrated zero approximation error for both metrics when the displacement from the sampling point was zero. This result confirms that the coresets precisely recovered the original quadratic function at the sampling point. Although quadratic approximation also showed zero error at the sampling point, its accuracy quickly deteriorated as the evaluation point moved away from the sampling point. This deterioration suggests that the registration error factor is highly nonlinear and cannot be adequately captured by linear approximations. Random sampling exhibited large approximation errors with this very small number of sampled points. For example, random sampling with 256 points resulted in mean errors of 0.698 m and 0.132\textdegree and KLD errors of $2.7 \times 10^3$ and $4.1 \times 10^3$ on average. These results indicate that random sampling requires a significantly larger number of points (e.g., more than 1000) to reasonably approximate the original error function.

\subsection{Robustness to Point Cloud Degeneration}
\label{sec:ex_flatwall}

\begin{figure}[tb]
  \centering
  \includegraphics[width=0.55\linewidth]{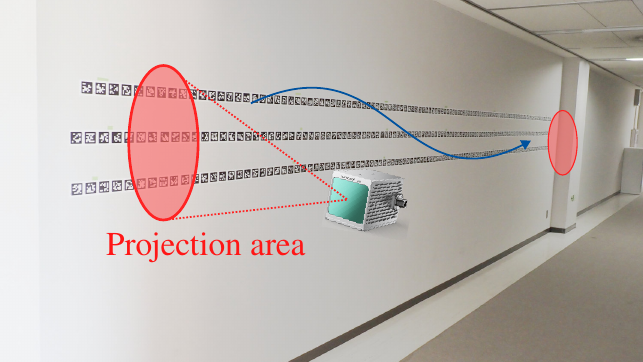}
  \caption{Setup for flat wall experiment~\cite{Koide_2024}. The LiDAR was moved between pillars, which induced severe point cloud degeneration for a few seconds.}
  \label{fig:flatwall}
\end{figure}

\begin{table}[tb]
  \centering
  \caption{Point Cloud Degeneration Test Results (ATEs [m])}
  \label{tab:degenerated_real}
  \scriptsize
  \begin{tabular}{g|ggggg}
  \toprule
  \rowcolor{white}
  \rowcolor{white}
  Seq.    & FLIO \cite{Xu2022} & VoxelMap \cite{Yuan_2022} & SLICT \cite{Nguyen_2023} & GLIM \cite{Koide_2024} & Proposed       \\
  \midrule
  \midrule
  \rowcolor{white}
  01      & 0.815   & 0.577  & 0.947  & 0.118  & 0.424 \\
  02      & 0.822   & 0.146  & 0.331  & 0.299  & 0.092 \\
  \rowcolor{white}
  03      & 0.873   & 0.950  & 1.088  & 0.040  & 0.114 \\
  04      & 1.137   & 0.586  & 0.729  & 0.389  & 0.448 \\
  \rowcolor{white}
  05      & 1.048   & 0.786  & 0.747  & 0.228  & 0.311 \\
  06      & 15.551  & 0.807  & 0.311  & 0.056  & 0.068 \\
  \rowcolor{white}
  07      & 0.635 & 0.366  & 0.659  & 0.017  & 0.014 \\
  08      & 0.297 & 0.279  & 0.983  & 0.146  & 0.045 \\ 
  \midrule
  \rowcolor{white}
  Average & 2.647 & 0.562 & 0.724  & 0.162  & 0.190 \\
  \bottomrule
  \end{tabular}
\end{table}

{\bf Experimental setting: } We evaluated the proposed odometry estimation method on the {\it flatwall} dataset\footnote{\url{https://staff.aist.go.jp/k.koide/projects/glimsupp/flatwall.html}} to demonstrate its robustness to point cloud degeneration. This dataset consists of eight sequences recorded with a Livox Avia. In each sequence, the LiDAR was moved between two pillars in a corridor while facing a flat wall, which induced severe point cloud degeneration for a few seconds. While the durations of the sequences are relatively short (8.8 to 18.5 s), this dataset poses a significant challenge for existing methods, as point cloud degeneration makes accurate odometry estimation difficult.

We compared the proposed method with four state-of-the-art tightly coupled LiDAR-IMU odometry methods, namely FAST-LIO2 (FLIO) \cite{Xu2022}, based on the iterated Kalman filter, VoxelMap \cite{Yuan_2022}, based on a voxel-based efficient plane representation, SLICT \cite{Nguyen_2023}, which uses continuous-time sliding window optimization, and GLIM \cite{Koide_2024}, which uses GPU-accelerated sliding window optimization.

{\bf Experimental results: } Table \ref{tab:degenerated_real} summarizes the absolute trajectory errors (ATEs) \cite{Zhang_2018} for the evaluated methods. FAST-LIO2, VoxelMap, and SLICT were greatly affected by the severe degeneration of point clouds and thus exhibited large average ATEs (2.647, 0.562, and 0.724 m, respectively). The proposed method showed an average ATE of 0.190 m, which is comparable to that of GLIM (0.162 m), even though it does not rely on GPU acceleration. This performance is attributed to sliding window optimization, which effectively corrects trajectory drift during point cloud degeneration by propagating future observations to past states.

\subsection{Quantitative Evaluation on MCD VIRAL dataset}
\label{sec:ex_mcd}

{\bf Experimental setting: } To quantitatively evaluate the accuracy of the proposed method in practical scenarios, we conducted experiments on the MCD VIRAL dataset \cite{mcdviral2024}. This dataset consists of 18 sequences recorded with an Ouster OS1-128 and a Livox Mid70. Six sequences (ntu sequences) were recorded with an on-vehicle setup. The remaining sequences (kth and tuhh sequences) were recorded with a handheld sensor setup. The average and maximum durations of the sequences are respectively 537 and 969 s. 

\begin{table}[tb]
  \caption{ATEs [m] for Odometry Estimation Methods on MCD VIRAL Dataset}
  \label{tab:mcd_results_odom}
  \centering
  \scriptsize
  \begin{tabular}{g|gggggg}
  \toprule
  \rowcolor{white}
  Sequence          & FLIO \cite{Xu2022} & DLIO \cite{Chen_2023} & SLICT \cite{Nguyen_2023} & GLIM \cite{Koide_2024} & Proposed \\
  \midrule
  \midrule
  \rowcolor{white}
  ntu\_day\_01    &     1.510  & 1.925  &     1.890  &     1.054 & \bf 0.918  \\
  ntu\_day\_02    &     0.272  & 0.636  & \bf 0.168  &     0.259 &     0.289  \\
  \rowcolor{white}
  ntu\_day\_10    &     2.084  & 3.052  &     1.429  & \bf 1.099 &     1.361  \\
  ntu\_night\_04  &     1.599  & 2.373  &     1.002  &     1.007 & \bf 0.961  \\
  \rowcolor{white}
  ntu\_night\_08  &     1.425  & 2.056  & \bf 0.822  &     1.558 &     1.846  \\
  ntu\_night\_13  &     0.903  & 1.928  & \bf 0.574  &     0.785 &     0.780  \\
  \rowcolor{white}
  kth\_day\_06    &     1.005  & 0.562  &     0.633  & \bf 0.283 &     0.466  \\
  kth\_day\_09    &     0.733  & 0.326  &     0.262  & \bf 0.194 &      0.219  \\
  \rowcolor{white}
  kth\_day\_10    &     2.176  & 0.665  &     0.737  & \bf 0.204 &     0.296  \\
  kth\_night\_01  &     1.040  & 0.414  &     0.540  & \bf 0.317 &     0.403  \\
  \rowcolor{white}
  kth\_night\_04  &     0.567  & 0.376  &     0.441  & \bf 0.152 &     0.256  \\
  kth\_night\_05  &     2.158  & 0.903  &     0.855  &     0.259 & \bf 0.185  \\
  \rowcolor{white}
  tuhh\_day\_02   &     0.273  & 0.283  &     0.236  & \bf 0.185 &     0.268  \\
  tuhh\_day\_03   &     0.970  & 0.731  &     0.743  &     0.843 & \bf 0.459  \\
  \rowcolor{white}
  tuhh\_day\_04   & \bf 0.077  & 0.232  &     0.084  &     0.124 &     0.083  \\
  tuhh\_night\_07 &     0.279  & 0.436  &     0.227  & \bf 0.120 &     0.272  \\
  \rowcolor{white}
  tuhh\_night\_08 &     0.749  & 0.685  &     0.740  &     0.605 & \bf 0.520  \\
  tuhh\_night\_09 & \bf 0.057  & 0.375  &     0.094  &     0.089 &     0.067  \\
  \midrule
  \rowcolor{white}
  Average         &     0.993  & 0.998  &     0.638  &     0.508 &     0.536 \\
  \bottomrule
  \end{tabular}
  \\
  Best values are shown in bold.
\end{table}

{\bf Odometry estimation accuracy: } We compared the proposed odometry estimation with FAST-LIO2 \cite{Xu2022}, DLIO \cite{Chen_2023}, SLICT \cite{Nguyen_2023}, and GLIM \cite{Koide_2024}. Loop closure was disabled for all methods. Table \ref{tab:mcd_results_odom} summarizes the ATEs for the evaluated methods. The results for FAST-LIO2, DLIO, and SLICT were taken from \cite{mcdviral2024}. Note that our method used only point clouds obtained using Ouster OS1-128.

FAST-LIO2 and DLIO, based on state filtering, were greatly affected by the challenging setup, which involved high-speed motion and dynamic environment changes, and thus exhibited large average ATEs (0.993 and 0.998 m, respectively). Although SLICT exhibited a better ATE (0.638 m), it was computationally intensive and failed to achieve real-time processing. Among the compared methods, GLIM showed the best average ATE (0.508 m) owing to its robust sliding window optimization. However, it required GPU acceleration for real-time processing. The proposed method showed an average ATE (0.536 m) that was comparable to that of GLIM owing to sliding window optimization. 

\begin{table}[tb]
  \caption{ATEs [m] for Global Trajectory Optimization Methods with Loop Closure on MCD VIRAL Dataset}
  \label{tab:mcd_results_map}
  \centering
  \scriptsize
  \begin{tabular}{g|gggg}
  \toprule
  \rowcolor{white}
  Sequence         & SLICT \cite{Nguyen_2023} & PGO \cite{Koide_2024} & GLIM \cite{Koide_2024} & Proposed \\
  \midrule
  \midrule
  \rowcolor{white}
  ntu\_day\_01    &     0.885  &     0.853  & \bf 0.686 &     0.731 \\
  ntu\_day\_02    & \bf 0.179  &     0.390  &     0.185 &     0.255 \\
  \rowcolor{white}
  ntu\_day\_10    &     1.108  &     0.815  & \bf 0.625 &     0.689 \\
  ntu\_night\_04  &     1.312  &     0.956  &     0.983 & \bf 0.881 \\
  \rowcolor{white}
  ntu\_night\_08  &     0.678  &     0.678  &     0.582 & \bf 0.560 \\
  ntu\_night\_13  &     0.717  &     0.667  &     0.668 & \bf 0.619 \\
  \rowcolor{white}
  kth\_day\_06    &     0.330  &     0.364  & \bf 0.144 &     0.227 \\
  kth\_day\_09    &     0.122  &     0.169  &     0.109 & \bf 0.098 \\
  \rowcolor{white}
  kth\_day\_10    &     0.250  &     0.327  &     0.220 & \bf 0.174 \\
  kth\_night\_01  &     0.340  &     0.318  &     0.277 & \bf 0.248 \\
  \rowcolor{white}
  kth\_night\_04  &     0.200  &     0.292  &     0.128 & \bf 0.095 \\
  kth\_night\_05  &     0.329  & \bf 0.185  &     0.241 &     0.203 \\
  \rowcolor{white}
  tuhh\_day\_02   &     0.190  &     0.215  &     0.125 & \bf 0.119 \\
  tuhh\_day\_03   &     0.472  & \bf 0.360  &     0.718 &     0.593 \\
  \rowcolor{white}
  tuhh\_day\_04   &     0.069  &     0.111  &     0.069 & \bf 0.063 \\
  tuhh\_night\_07 &     0.199  &     0.165  & \bf 0.129 &     0.134 \\
  \rowcolor{white}
  tuhh\_night\_08 & \bf 0.403  &     0.512  &     0.552 &     0.500 \\
  tuhh\_night\_09 &     0.103  &     0.075  & \bf 0.055 &     0.058 \\
  \midrule
  \rowcolor{white}
  Average         &     0.438  &     0.414  &     0.361 &     0.347 \\
  \bottomrule
  \end{tabular}
  \\
  Best values are shown in bold.
\end{table}

\begin{figure}[tb]
  \centering
  \begin{minipage}[tb]{0.48\linewidth}
  \centering
  \includegraphics[width=\linewidth]{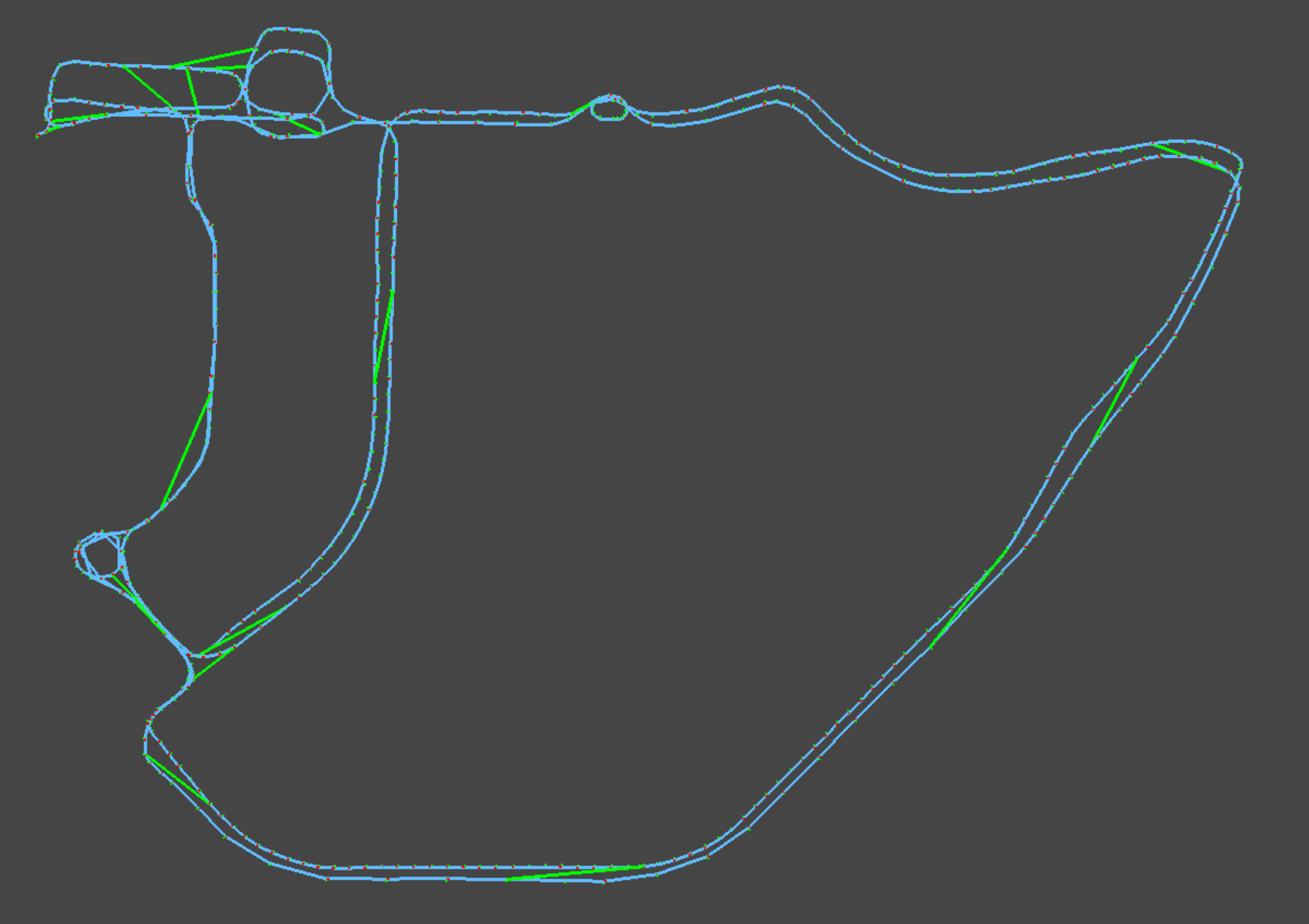}
  \subcaption{SLICT}
  \end{minipage}
  \begin{minipage}[tb]{0.48\linewidth}
  \centering
  \includegraphics[width=\linewidth]{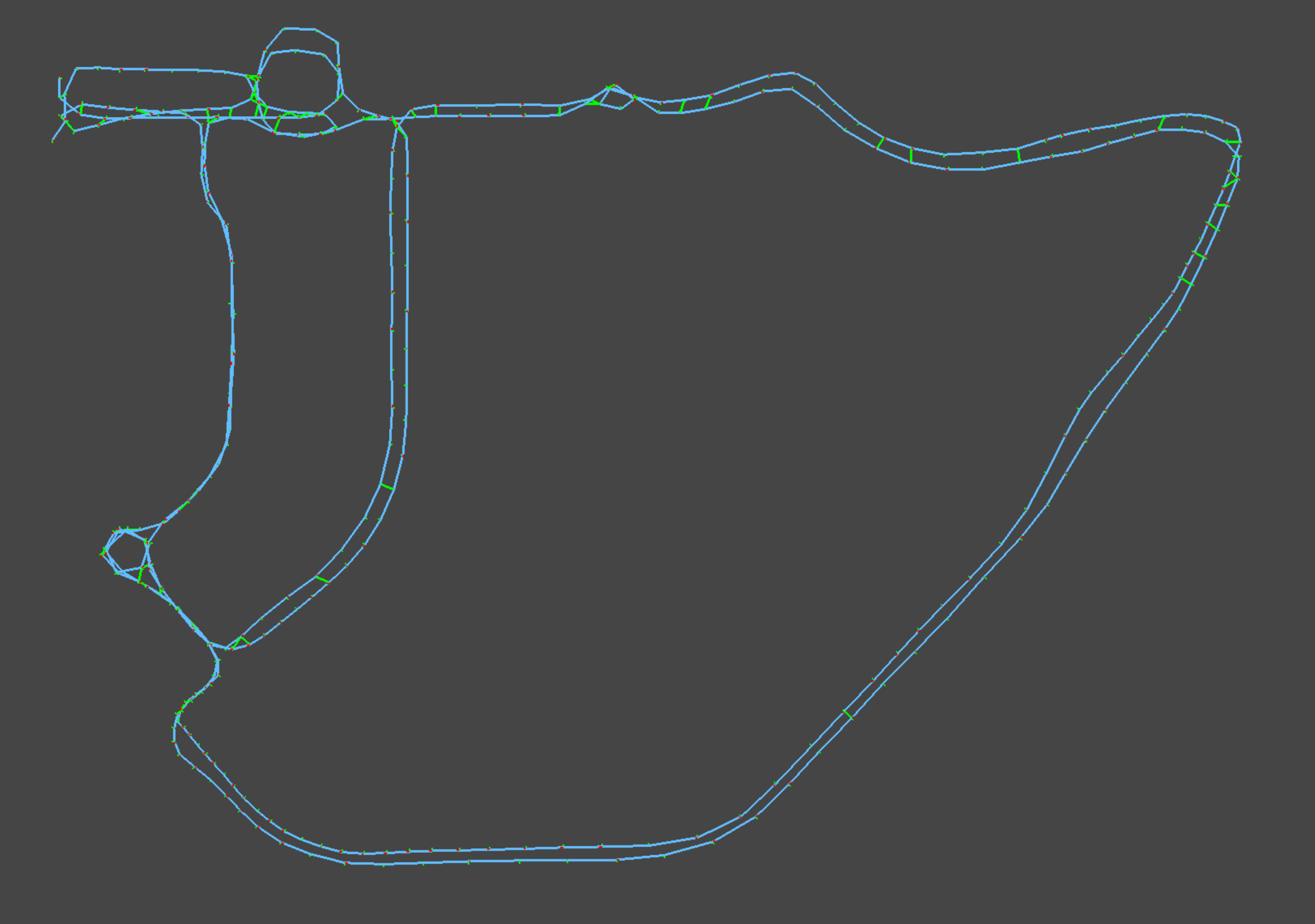}
  \subcaption{PGO}
  \end{minipage}
  \caption{Global optimization graphs. Blue and green lines respectively represent odometry and loop factors.}
  \label{fig:graphs_pgo}
\end{figure}

\begin{figure}[t]
  \centering
  \begin{minipage}[tb]{0.62\linewidth}
  \centering
  \includegraphics[height=4.6cm]{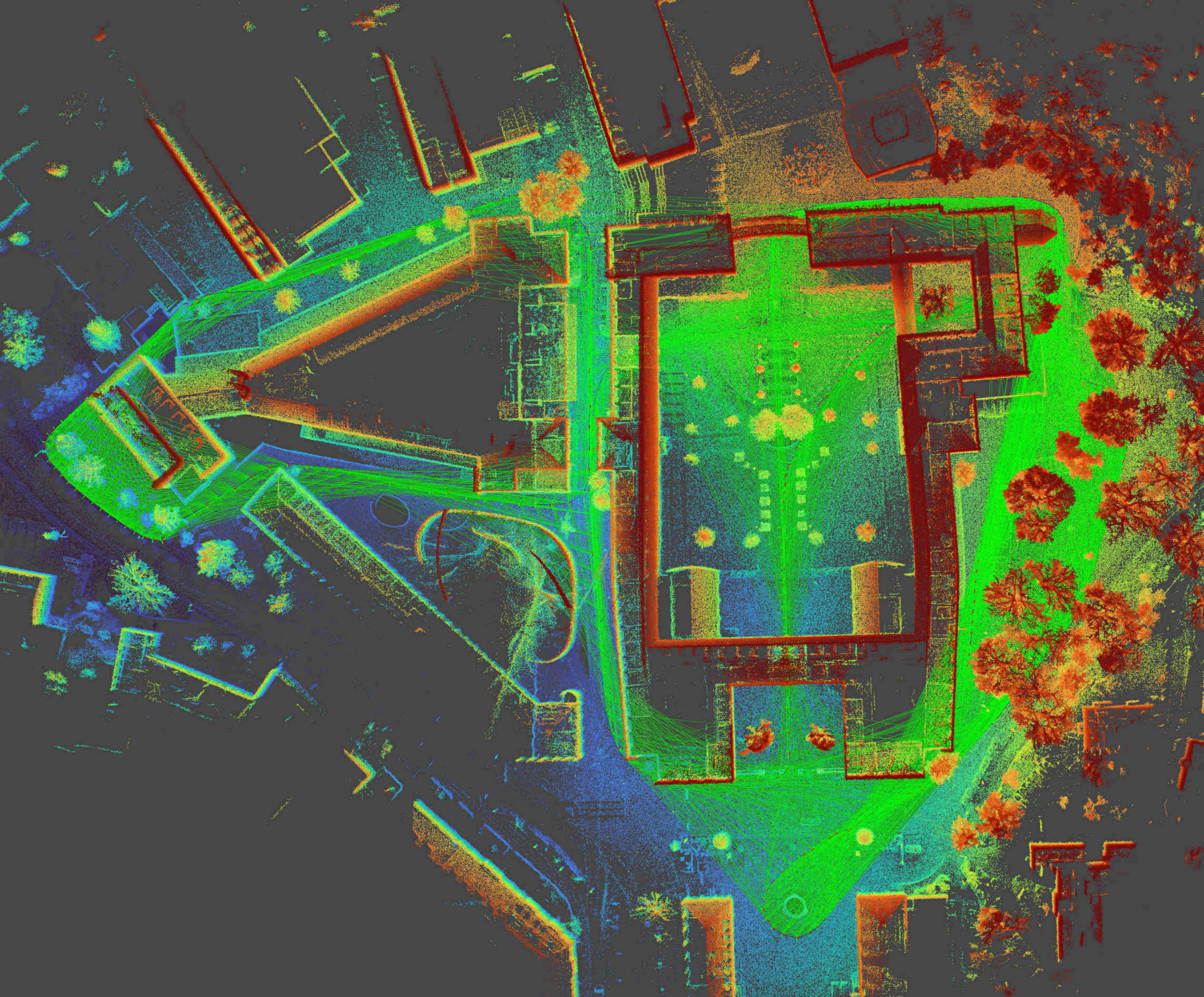}
  \subcaption{kth\_day\_06}
  \end{minipage}
  \begin{minipage}[tb]{0.35\linewidth}
  \centering
  \includegraphics[height=4.6cm]{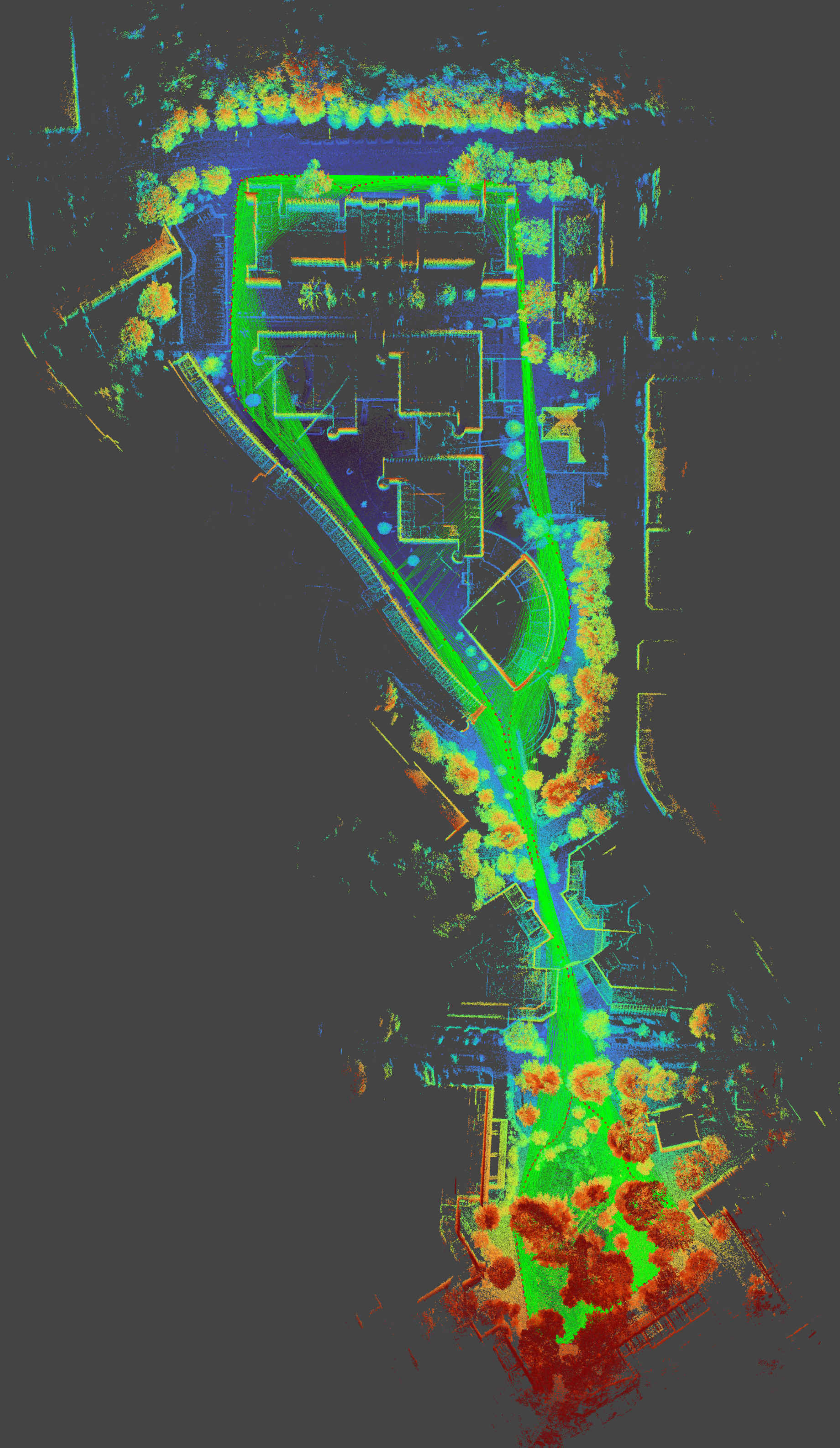}
  \subcaption{tuhh\_day\_03}
  \end{minipage}
  \caption{Example of mapping results and global trajectory optimization graphs.}
  \label{fig:graphs_proposed}
\end{figure}

{\bf Global mapping accuracy: } We compared the proposed SLAM framework including loop closure with three baseline methods. SLICT \cite{Nguyen_2023} uses conventional scan-matching-based loop detection and pose-graph-based trajectory optimization. PGO is conventional pose-graph-based trajectory optimization implemented in GLIM \cite{Koide_2024}, similar to that of SLICT. GLIM \cite{Koide_2024} uses the same global trajectory optimization factor graph structure as that of the proposed method with GPU-accelerated registration error factors.

Table \ref{tab:mcd_results_map} summarizes the ATEs of the compared methods with loop closure. Although SLICT had a greatly improved accuracy compared to that without loop closure, it showed the largest ATE among the compared methods (0.438 m). PGO showed the second largest average ATE (0.414 m). In particular, SLICT and PGO exhibited large ATEs in longer sequences (ntu sequences). Fig.~\ref{fig:graphs_pgo} shows the pose graphs generated by SLICT and PGO for the ntu\_day\_01 sequence. Blue and green lines represent odometry and loop constraints, respectively. The generated pose graphs are very sparse, as pose-graph-based methods have difficulty closing loops for frames with a small overlap. This result suggests that it is difficult to accurately correct trajectories with large loops using pose graph optimization.

GLIM and the proposed method showed better ATEs compared to those of the pose-graph-based methods (0.361 and 0.347 m, respectively). Fig.~\ref{fig:graphs} (b) and Fig.~\ref{fig:graphs_proposed} show that the proposed method constructed dense registration error minimization factor graphs that enabled accurate correction of trajectories with large loops. The proposed method exhibited a slightly better average ATE compared to that of GLIM. Although the difference is not significant, we consider that the use of exact nearest neighbor search resulted in better convergence compared to that for GLIM, which is based on voxel-based approximate nearest neighbor search.

%

\begin{figure}[tb]
  \centering
  \includegraphics[width=\linewidth]{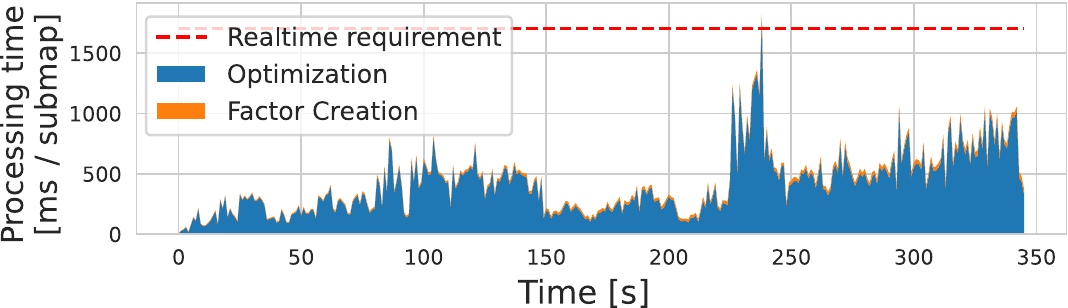}
  \caption{Processing time of global mapping for ntu\_day\_01.}
  \label{fig:time}
\end{figure}

{\bf Processing time: } We measured the processing times of the odometry estimation and the global mapping modules on the ntu\_day\_01 sequence, one of the longest sequences in the dataset. The timings were recorded on an Intel Core i7 8700K with 32 GB of RAM. Note that the factor linearization and the linear solver were made multi-threaded using Intel TBB.

The odometry estimation and the global mapping took approximately 51.9 ms (for each frame) and 433.9 ms (for each submap), respectively, both well within the real-time requirements (100 ms per frame and an average submap interval of 1.7 s). Fig.~\ref{fig:time} shows how the processing time of the global mapping module changed as the mapping progressed. Thanks to the exact downsampling algorithm and incremental optimization with iSAM2, the global optimization quickly converged every time a new submap was inserted. 

As an ablation study, we measured the processing times with exact downsampling disabled.
The average processing times of odometry estimation and global trajectory optimization deteriorated significantly, falling below the real-time requirements (284.8 and 1831.8 ms, respectively). This result confirms that exact downsampling greatly accelerates the optimization process.


\section{Conclusions}

We proposed a range-inertial SLAM framework with GICP factors accelerated with the exact downsampling algorithm. We devised a deferred sampling strategy to mitigate the processing cost of exact downsampling. The experimental results demonstrated that the proposed method enables the real-time processing of registration error minimization factor graphs on a standard CPU and exhibits an accuracy comparable to those of GPU-based methods.

\balance

\bibliographystyle{IEEEtran}
\bibliography{icra2025}

\end{document}